\def\year{2021}\relax
\documentclass[letterpaper]{article} 
\usepackage{aaai21}  
\usepackage{times}  
\usepackage{helvet} 
\usepackage{courier}  
\usepackage[hyphens]{url}  
\usepackage{graphicx} 
\usepackage{natbib}  
\usepackage{caption} 
\usepackage{comment}
\usepackage{times}
\usepackage{latexsym}
\usepackage{amsfonts}
\usepackage{graphicx}
\usepackage{caption}
\usepackage{subcaption}
\usepackage{url}
\usepackage{bm}

\usepackage{booktabs}
\usepackage{amssymb}
\usepackage{amsmath}
\usepackage{enumitem}
\usepackage{mathtools}
\title{Fair Hate Speech Detection through Evaluation of Social Group Counterfactuals }
\author{
    Aida Mostafazadeh Davani\textsuperscript{\rm 1},
    Ali Omrani\textsuperscript{\rm 1},
    Brendan Kennedy\textsuperscript{\rm 1},
    Mohammad Atari\textsuperscript{\rm 2},
    Xiang Ren\textsuperscript{\rm 1},
    Morteza Dehghani\textsuperscript{\rm 1}\textsuperscript{\rm 2}\\
}
\affiliations{
    \textsuperscript{\rm 1}Department of Computer Science\\ University of Southern California\\
    \textsuperscript{\rm 2}Department of Psychology\\ University of Southern California\\
    mostafaz@usc.edu,
    aomrani@usc.edu,
    btkenned@usc.edu,
    atari@usc.edu,
    xiangren@usc.edu,
    mdehghan@usc.edu

}

\begin{document}

\maketitle

\begin{abstract}
Approaches for mitigating bias in supervised models are designed to reduce models' dependence on specific sensitive features of the input data, e.g., mentioned social groups. 
However, in the case of hate speech detection, it is not always desirable to equalize the effects of social groups because of their essential role in distinguishing outgroup-derogatory hate, such that particular types of hateful rhetoric carry the intended meaning only when contextualized around certain social group tokens.
\textit{Counterfactual token fairness} for a mentioned social group evaluates the model's predictions as to whether they are the same for (a) the actual sentence and (b) a \textit{counterfactual} instance, which is generated by changing the mentioned social group in the sentence. Our approach assures robust model predictions for counterfactuals that imply similar meaning as the actual sentence. To quantify the similarity of a sentence and its counterfactual, we compare their likelihood score calculated by generative language models. By equalizing model behaviors on each sentence and its counterfactuals, we mitigate bias in the proposed model while preserving the overall classification performance. 
\end{abstract}

\section{Introduction}
Hate speech classifiers have high false-positive error rates on documents that contain specific social group tokens (SGTs; e.g., Asian, Jew), due in part to the high prevalence of SGTs in instances of hate speech \citep{wiegand2019detection, mehrabi2019survey}. This \textit{unintended bias} \citep{dixon2018measuring} is illustrated by the high frequency of ``Muslim'', for example, in hate speech-related instances of the train set, and the consequent higher false-positive errors for posts that include the word ``Muslim''.

Several existing frameworks offer methods for countering unintended bias based on counterfactual fairness. Counterfactual fairness considers the change in model prediction in a counterfactual situation by changing the SGT in the input. Fairness evaluation metrics, e.g., \textit{equality of odds} and \textit{equality of opportunity}, require model predictions to be \textit{robust} in counterfactual situations \citep{hardt2016equality}.
Satisfying these metrics has motivated data augmentation approaches to balance the data distribution \citep{dixon2018measuring, zhao2018gender, park2018reducing} or fair input representations to equalize model performance with respect to protected groups \citep{madras2018learning, zhang2018mitigating}. %

However, when SGTs have a definitional contribution to the semantics of a construct, as they do for hate speech, 
the contribution of the SGT to preserving the meaning of the sentence should be concomitantly considered prior to expecting robust model predictions for counterfactuals \citep{haas2012hate}. In fact, social groups are mentioned in specific contexts based on how they are socially perceived and stereotyped \citep{fiske2002model,warner2012detecting}. For instance, if a document includes a stereotype about Muslims (e.g., calling a Muslim terrorist because of their religion), changing the word ``Muslim'' to ``Jew'' underscores the interaction of the context and the SGT since the same stereotypes do not hold and are not usually used for Jews.
Therefore, robust model behaviors should be restricted to counterfactuals that are similar to the actual sentence.

\begin{figure}[t]
    \centering
    \includegraphics[width=\linewidth]{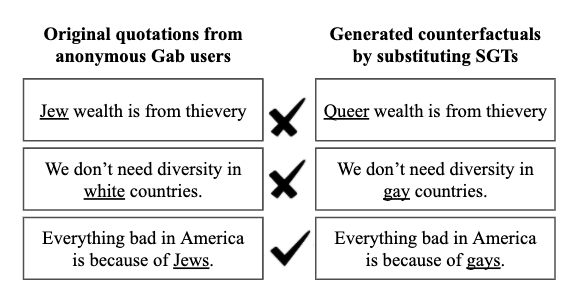}
    \caption{Two examples of asymmetric counterfactuals, and one example of a symmetric counterfactual.}
    \label{fig:counterfactual}
\end{figure}

In this paper, rather than equalizing model behavior for all counterfactuals of a sentence, 
we restrict counterfactual reasoning to cases where substituting the SGT conveys a similar meaning (e.g., we may not consider substituting ``Muslim'' with ``Jew'' in a hateful sentence about terrorism). In doing so, we detect and discard \textit{asymmetric counterfactuals}, in which the SGT substitution modifies the meaning of the text drastically (Figure \ref{fig:counterfactual}). To operationalize the meaning modification, we evaluate the decrease in the likelihood of the sentence, calculated by a pre-trained language model, as a result of counterfactual SGT substitution. 
During training, we equalize the classifier's predictions on sentences and their similar counterfactuals (\textit{symmetric counterfactuals}) by employing a logit pairing approach \citep{kannan2018adversarial}. 
We show that assuring similar performance on sentences and their symmetric counterfactuals helps pursue counterfactual token fairness \citep{garg2019counterfactual}. 

Our contributions are (1) proposing a method for excluding asymmetric counterfactual based on sentence likelihoods;
and (2) achieving fair predictions for social group pairs, based on their contextualized similarities. To this end, we first demonstrate the power of sentence likelihoods, calculated by a generative model, in distinguishing the association of sentences with their mentioned SGTs. We explore documents in a dataset of social media posts each mentioning exactly one social group and show that sentences differ based on whether they are predictive of their mentioned social group. Our results show that in a subset of the dataset, the SGT can exclusively be predicted using the likelihood of the sentence. Then, to apply counterfactual token fairness to hate speech detection, we use sentence likelihoods to differentiate SGTs that can interchangeably appear in a sentence. For each instance of the dataset, we apply counterfactual logit pairing using SGTs that result in the least amount of change in the meaning. Our experiments on two datasets show that our method can better improve fairness, while preserving classification performance, compared to other bias mitigation models.


\section{Related Work}


Hate speech detection models have been studied in fairness research in machine learning, given their biases towards SGTs. 
\citet{dixon2018measuring} defined unintended bias as differing performance on subsets of the dataset that contain particular SGTs.
When biased datasets initiate this issue, approaches for data augmentation are proposed to create a balanced ratio of positive and negative labels for each SGT or prevent biases from propagating to the learned model \citep{dixon2018measuring, zhao2018gender, park2018reducing}.

Other approaches modify training objectives via L2 norms of feature importance \citep{liu2019incorporating} or via regularization of post-hoc term-level importance \citep{kennedy2020posthoc}. Others apply adversarial learning for generating fair representations \citep{madras2018learning,zhang2018mitigating}
by minimizing predictability of preserved features from input data while maximizing classification accuracy. While fair representations have been applied in different machine learning problems to protect preserved attributes, \citet{elazar2018adversarial} demonstrated that adversarial learning cannot achieve invariant representation of features.

By altering sensitive features of the input and evaluating the changes in the output, counterfactual fairness \cite{kusner2017counterfactual} assesses the bias in machine learning models. Similarly, counterfactual token fairness defines a fair (i.e., unbiased) model as one that behaves consistently across counterfactual sets of instances \citep{garg2019counterfactual}.

\section{Dataset}
\label{sec:data}
In the present studies, we explore hate speech in a corpus of social media posts from Gab\footnote{\url{https://gab.com}}. We downloaded Gab posts from the public dump of the data by Pushshift.io\footnote{\url{https://files.pushshift.io/gab/}} \citep{pushshift_gab}. In the first study, we randomly selected 15 million posts from the Gab corpus, posted from August 2016 to October 2018. We analyze a subset of this dataset; \textbf{SGT-Gab}, which includes all posts that mention one SGT (\textit{N} = 2M).
In the second study, to train hate speech detection models, we used Gab Hate Corpus \citep[\textbf{GHC};][]{kennedy2020gab} and Stromfront dataset \citep[\textbf{Storm};][]{de2018hate}; including 27k and 11k social media posts respectively, annotated based on their hate speech content.

The list of SGTs (see Supplementary Materials) is compiled from \citet{dixon2018measuring} and extended using a Natural Language ToolKit \citep[NLTK;][]{loper2002nltk} function for WordNet synset generation. The resulting list includes 77 specific social group terms.

\section{Analysis of Context-SGT Interaction}
\label{sec:context}

As stated by \citet{warner2012detecting}, hate speech can include language that is offensive to any social group -- e.g., call for violence against a group \citep{kennedy2020gab} -- or prejudicial expressions which target individuals and groups based on their social stereotypes \cite{fiske2002model}. Therefore, any attempt for supporting social group fairness in hate speech detection (e.g., counterfactual fairness) requires essential considerations for stereotypical language that is exclusive to particular target groups. In such cases, expecting robust model performance for all counterfactuals of the sentence is not 
in accordance with fairness objectives.
Here, to indicating the extent of stereotypical language in the text, we identify a subset of a corpus of social media posts, in which SGTs can be predicted from their surrounding words.


We apply generic language models to evaluate the predictability of a mentioned SGTs among possible counterfactuals --- e.g., we expect the language model to predict a higher likelihood for a sentence about terrorism when it is paired with ``Muslim'' versus other SGTs. By doing so, we identify sentences that are significantly different from their counterfactuals. \citet{nadeem2020stereoset} show that generative models (e.g., GPT-2) exhibit strong stereotypical biases and therefore, perform well in detecting stereotype content. 
In this study, we consider all instances of \textbf{SGT-Gab} 
and construct counterfactuals through the substitution of SGTs. 

For an instance $\bm{x}$ in \textbf{SGT-Gab} with an SGT, $s_i$, and a set of possible SGTs $S$, the set of all counterfactual $\Theta(x)$ is: $$\{\text{substitute}(s_i, s_j)| \forall s_j \in S, j \neq i\}$$

\begin{figure}[b!]
    \centering
    \includegraphics[width=.47\textwidth]{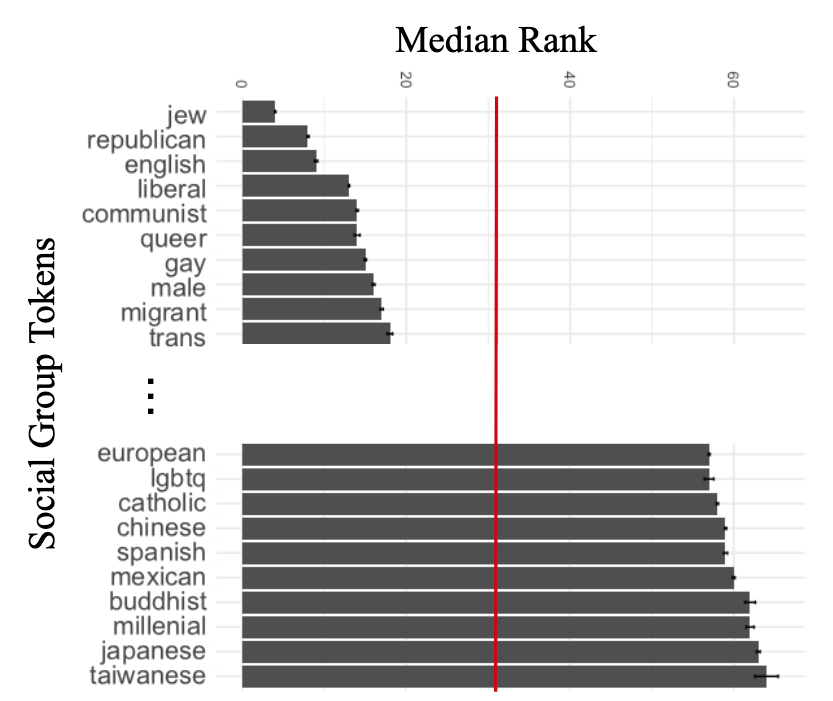}
    \caption{\textbf{The median log-likelihood ranks for sentences mentioning each social group}. Each sentence was ranked among its 64 counterfactuals, generated by altering the SGT. Stereotype-related language and contextual predictability vary significantly among SGTs}
    \label{fig:study1}
\end{figure}

To measure SGT predictability across contexts, for a given $\bm{x}$ and its counterfactual set $\Theta(\bm{x})$ we compute the likelihood assigned by a pre-trained language model, specifically GPT-2 \citep{radford2019language}. Notably, GPT-2 has  achieved high performance on detecting stereotypes in language \citep{nadeem2020stereoset} and is therefore suitable as a language representations model that embeds stereotypical relations at the sentence level. 

For each word $x_i$ in a sentence, the likelihood of $x_i$, $P(x_i|x_0\ldots x_{i-1})$, is approximated by the softmax of $x_i$ with respect to the vocabulary. Therefore, the log-likelihood of a sentence $x_0, x_1, \ldots x_{n-1}$ is computed with:
\begin{equation*}
  \lg P(\bm{x}) = \sum_{i}^n \lg P(x_i|x_0,..,x_{i-1})
\end{equation*}

The log-likelihood of each instance and its counterfactuals were computed for \textbf{SGT-Gab}. 
The primary outcome was the original instance's \textit{rank} in log-likelihood amongst its counterfactuals. Higher rank for a mentioned SGT implies a higher dependence on context and indicates the stereotypical content of the sentence.
In \textbf{SGT-Gab}, the aggregated results show that in 2.9\% of the sentences, the mentioned SGT achieves the best ranking and in 13.9\% of all posts, the mentioned SGT appears in the first 10\% rankings. 

Moreover, in stereotypical posts (in which the mentioned SGT achieves a high rank), we analyze whether highly ranked SGTs were conceptually related to the mentioned ones by comparing their associated social categories -- e.g., race, ethnicity, nationality, and gender. In 86.03\% of the posts, where the original SGT is ranked second, the top-ranked SGT is from the same social category. When the original SGT is ranked in the top 10\%, 72.46\% of SGTs with better ranking are from the same social categories. The results show that the similarity can be in-part explained by SGTs' common social category. This similarity can be further explored by quantifying social stereotypes regarding different social groups \citep{fiske2002model}.
Figure \ref{fig:study1} shows the averaged ranking of each SGT among all posts it appears in. The results show a high variation in averaged ranking among SGTs ($sd = 15.33$), indicating the variation of stereotypical content about each social group in the corpus.

\section{Asymmetric Counterfactual Filtering}
Designing approaches for satisfying the fairness criteria in hate speech classifier models requires specific steps for handling social group biases inherent in stereotyped language. 
However, we can infer from the results of our first analysis that in stereotype-related settings, substituting the SGT with other tokens potentially creates a counterfactual which should not be constrained to generate the same prediction (as the meaning of the instance has changed). These cases are referred to as asymmetric counterfactuals; here, we propose a method to detect them based on the change in sentence likelihood and ignore them during bias mitigation.


\subsection{Method}
We apply counterfactual logit pairing (CLP) to labeled instances and their counterfactuals \citep{garg2019counterfactual}. CLP penalizes divergence in output among a given input and its counterfactuals. 
Rather than simplifying the training process by exclusively training the logit pairing on all counterfactuals of negative instances of hate \citep{garg2019counterfactual}, we provide a procedure to identify asymmetric counterfactuals over the entire corpus.


We identify (and filter) counterfactuals based on their likelihood compared to that of the original sentence, calculated by GPT-2. 
Given a sentence $\bm{x}$, which includes an SGT, we generate the set of counterfactuals $\bm{x}_{cf}$ with higher log-likelihoods compared with $\bm{x}$:
\begin{equation*}
    x_{cf} = \{x | x' \in \Theta(x), P(x) \leq P(x') \}
\end{equation*}

\begin{table*}[]
    \centering
    \scalebox{0.9}{
    \begin{tabular}{l|cccc|cc|cc}
    \toprule
        \textbf{Model} & \multicolumn{4}{c|}{\textbf{Predicting Hate}} & \multicolumn{2}{c|}{\textbf{Equality of odds}} & \multicolumn{2}{c}{\textbf{CTF}}  \\
        & Acc & Precision & Recall & F1 & TP & TN & ASYM & SYM \\\midrule
        BiLSTM & 85.67 & \textbf{45.74} & 64.39& \textbf{53.38}& 21.25$\pm{32.4}$& 81.22$\pm{26.8}$& 0.50& 0.43 \\
        BiLSTM+Mask & \textbf{84.73} & 41.28 & 66.09 & 50.63 & 17.82$\pm{33.9}$& 81.61$\pm{\textbf{24.3}}$& 0.05 & 0.08 \\
        CLP+NEG & 84.19 & 36.26& 66.84& 46.76& 34.07$\pm{37.9}$ & 79.32$\pm{31.1}$ & 0.05 & 0.07 \\
        CLP+SC & 83.48 & 36.19 & \textbf{67.05} & 46.83 & 25.30$\pm{\textbf{27.8}}$ & 67.50$\pm{40.8}$ & 0.07 & 0.09 \\
        CLP+ASY & 84.38 & 41.50 & 63.02 & 49.60 & 21.80$\pm{31.5}$& 82.22$\pm{26.4}$& \textbf{0.04}& \textbf{0.04}\\
        \bottomrule
    \end{tabular}
    }
    \caption{Vanilla BiLSTM model, BiLSTM model with masking SGTs, baseline CLP, CLP method based on social categories, and our CLP method with asymmetric detection, trained in 5-fold cross validation and tested on 20\% of \textbf{Storm}. Fairness evaluations include true positive (TP) and true negative (TN) as the metrics for equality of odds, and counterfactual token fairness (CTF) over two datasets of counterfactuals.}
    \label{tab:results2}
\end{table*}

Consequently, semantically different counterfactuals are not considered for mitigating bias in stereotypical content.
Given the generated set of counterfactuals, $\bm{x}_{cf}$, a classifier, $f$, satisfies counterfactual fairness if \citep{garg2019counterfactual}:
\begin{equation*}
    |f(x) - f(x'))| < \epsilon, \forall x \in X, \forall x' \in x_{cf}
\end{equation*}
where $X$ contains the whole annotated dataset. The averaged $|f(x) - f(x'))|$ among the predictions of a model is considered as the measurement of \textit{counterfactual token fairness} (CTF).

\subsection{Experiment and Results}
We compare three BiLSTM \citep{schuster1997bidirectional} classifiers with CLP, one trained based on our approach for excluding asymmetric counterfactuals (CLP+ASY), one based on \citet{garg2019counterfactual}'s counterfactual generation (CLP+NEG), which attempts to remove asymmetric counterfactuals by considering all counterfactuals for negative instances of hate. To evaluate alternative strategies for CLP, the third model generates counterfactuals based on social categories (CLP+SC). E.g., for a sentence mentioning a racial group, we only consider counterfactuals that mention racial groups. We also compare the accuracy and fairness of these models with a BiLSTM baseline with no bias mitigation (BiLSTM), and a BiLSTM model that masks the SGTs (BiLSTM+Mask). Table \ref{tab:results2} shows the results of analyzing these five models on \textbf{Storm} dataset. The results of comparing these methods on \textbf{GHC} is included in the Appendix.


Once a model achieves baseline accuracy scores on the hate recognition task, fairness scores are reported based on fairness criteria.
First, we evaluate the measurements for equality of odds. Namely, we compare the averaged rate of true positive (TP) and true negative (TN) results for predicting the hate speech label associated with each SGT in the preserved test set (20\% of the dataset). We then compute counterfactual token fairness (CTP) -- averaged $|f(x) - f(x'))|$ for sentences and their counterfactuals -- for two datasets of symmetric counterfactuals \citep{dixon2018measuring} and asymmetric counterfactuals \citep{nadeem2020stereoset}.

Symmetric counterfactuals (SYM) from \citet{dixon2018measuring} include synthetic instances based on templates (\textless You are a \verb|ADJ| \verb|SGT|\textgreater{}, and \textless Being \verb|SGT| is \verb|ADJ|\textgreater{}). In such instances, the context is explicitly dis-aggregated from the SGT, and the model prediction should solely depend on the \verb|ADJ|s.
Therefore, we expect smaller values of CTF for fair models. Asymmetric counterfactuals (ASYM) from \citet{nadeem2020stereoset} include stereotypical sentences and their counterfactuals which we generated by substituting the SGTs. Since all these instances are stereotypical, we expect all counterfactuals to be asymmetric, and CTF to be higher for this dataset.

\subsection{Discussion}

We demonstrate the stereotypical content of a sentence by comparing its log-likelihood to those of its counterfactuals, using a generative language model. This improves the approach for counterfactual token fairness \citep[CTF;][]{garg2019counterfactual} as it models the interaction of a mentioned social group with its sentence, and suggests further explorations of how a fair model should treat social groups equally based on the context. Experiments showed that discarding asymmetric counterfactuals -- which fail to convey the likelihood of the sentence -- improves CTF while seeing minimal changes in hate speech detection performance. Our model did not increase CTF for asymmetric samples which is in part due to the fact that stereotypical content is not always hate-related.

\section{Conclusion}
We show that the textual context can be variably associated with the social groups they mention. While stereotypical sentences include semantic clues of what social group they mention, other sentences imply the same meaning when paired with different social group tokens. We used this information to apply counterfactual reasoning for evaluating models' robust predictions upon a change in the social group token. Our method treats social groups equally according to the context, by applying logit pairing on a restricted set of counterfactuals for each instance. By doing so, counterfactual token fairness improved while the general accuracy and other fairness metrics were maintained.
Future work will explore alternative techniques for measuring asymmetry in social group counterfactuals and other domains for which our methods can be applied. By considering asymmetric counterfactuals in the method, we can formally model social group differentiation along with similarities, which can shed light on the textual associations of hate speech and stereotype.

\bibliography{ref}
\end{document}